\documentclass{article}

\usepackage{times}
\usepackage{graphicx} 
\usepackage{subfigure} 

\usepackage{algorithm}
\usepackage{algorithmic}

\usepackage{hyperref}

\usepackage[accepted]{icml2017} 

\usepackage{dirtytalk}
\usepackage{pdfpages}
\usepackage{float}
\usepackage{amsmath}
\usepackage{amssymb}

\icmltitlerunning{A step towards procedural terrain generation with GANs}

\begin{document} 

\twocolumn[
\icmltitle{A step towards procedural terrain generation with GANs}



\icmlsetsymbol{equal}{*}

\begin{icmlauthorlist}
\icmlauthor{Christopher Beckham}{mila}
\icmlauthor{Christopher Pal}{mila}
\end{icmlauthorlist}

\icmlaffiliation{mila}{Montr\'eal Institute of Learning Algorithms, Qu\'{e}bec, Canada}

\icmlcorrespondingauthor{Christopher Beckham}{christopher.beckham@polymtl.ca}

\icmlkeywords{ordinal, unimodal, kappa, neural networks, deep learning, machine learning, ICML}

\vskip 0.3in
]



\printAffiliationsAndNotice{}  

\begin{abstract} 
Procedural terrain generation for video games has been traditionally been done with smartly designed but handcrafted algorithms that generate heightmaps. We propose a first step toward the learning and synthesis of these using recent advances in deep generative modelling with openly available satellite imagery from NASA. 
\end{abstract} 

\section{Introduction}

Procedural generation in video games is the algorithmic generation of content intended to increase replay value through interleaving the gameplay with elements of unpredictability. This is in contrast to the more traditional, `handcrafty' generation of content, which is generally of higher quality but with the added expense of labour. A prominent game whose premise is almost entirely based on procedural terrain generation is Minecraft, a game where the player can explore a vast open world whose terrain is based entirely on voxels (`volumetric pixels'), allowing the player to manipulate the terrain (i.e. dig tunnels, build walls) and explore interesting landscapes (i.e. beaches, jungles, caves).

So far, terrains have been procedurally generated through a host of algorithms designed to mimic real-life terrain. Some prominent examples of this include Perlin noise \cite{perlin_noise} and diamond square \cite{fournier}, in which a greyscale image is generated from a noise source (with intensities proportional to heights above sea level), which, when rendered in 3D as a mesh, produces a terrain. While these methods are quite fast, they generate terrains which are quite simple in nature. Software such as L3DT employ sophisticated algorithms which let the user control what kind of terrain they desire, (e.g. mountains, lakes, valleys), and while these can produce very impressive terrains \footnote{See \url{http://www.bundysoft.com/docs/doku.php?id=l3dt:algorithms}}, it would still seem like an exciting endeavour to leverage the power of generative networks in deep learning (such as the GAN \cite{gan}) to learn algorithms to automatically generate terrain, without the need to manually write algorithms to do so.

\begin{figure*}
    \centering
    \subfigure[World heightmap]{\label{fig:earth_heightmap}\includegraphics[width=0.4\textwidth]{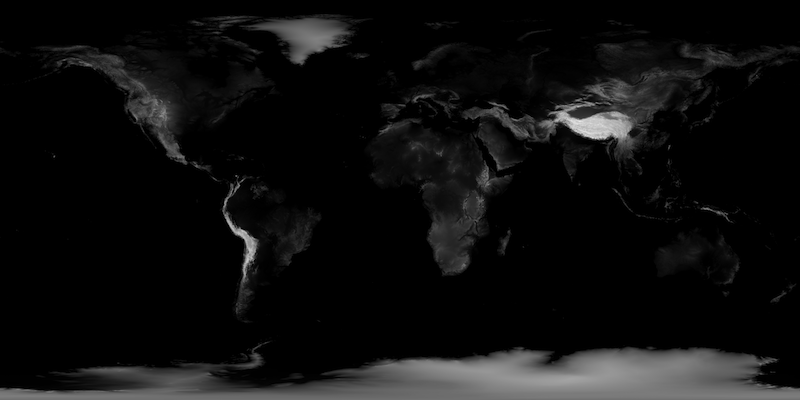}} \subfigure[World texture map]{\label{fig:earth_texture}\includegraphics[width=0.4\textwidth]{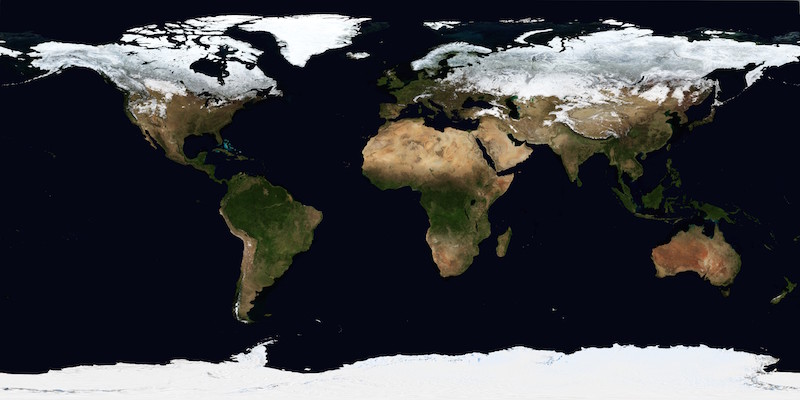}} 
    \caption{Heightmap and texture map (21600px x 10800px) of the earth provided by the NASA Visible Earth project. Both maps provide a spatial resolution of 1 square km per pixel.}
    \label{fig:earth}
\end{figure*}

In this paper, we leverage extremely high-resolution terrain and heightmap data provided by the NASA `Visible Earth' project\footnote{\url{https://visibleearth.nasa.gov/}} in conjunction with generative adversarial networks (GANs) to create a two-stage pipeline in which heightmaps can be randomly generated as well as a texture map that is inferred from the heightmap. Concretely, we synthesise 512px height and texture maps using random 512px crops from the original NASA images (of size 21600px x 10800px), as seen in Figure \ref{fig:earth}.

\subsection{Formulation}

Suppose $\mathbf{z}' \sim p(\mathbf{z})$ is a $k$-dimensional sample we draw from the prior distribution, $\mathbf{x}' = G_{h}(\mathbf{z}')$ the heightmap which is generated from $\mathbf{z}'$, and $\mathbf{y}' = G_{t}(\mathbf{x}')$ is the texture generated from the corresponding heightmap. We can think of this process as being comprised of two GANs: the `DCGAN' \cite{dcgan} which generates the heightmap from noise, and `pix2pix' \cite{pix2pix}, which (informally) refers to conditional GANs for image-to-image translation. If we denote the DCGAN generator and discriminator as $G_{h}(\cdot)$ and  $D_{h}(\cdot)$ respectively (where the `h' in the subscript denotes `heightmap'), then we can formulate the training objective as:
\begin{equation}
\begin{split}
\min_{G_{h}} \ & \ \ell( D_{h}(G_{h}(\mathbf{z}')), 1 ) \\
\min_{D_{h}} \ & \ \ell( D_{h}(\mathbf{x}), 1 ) + \ell( D_{h}(G_{h}(\mathbf{z}')), 0 ),
\end{split}
\label{eq:1}
\end{equation}
where $\ell$ is a GAN-specific loss, e.g. binary cross-entropy for the regular GAN formulation, and squared error for LSGAN \cite{lsgan}. We can write similar equations for the pix2pix GAN, where we now have $G_{t}(\cdot)$ and $D_{t}(\cdot, \cdot)$ instead (where `t' denotes `texture'), and instead of $\mathbf{x}$ and $\mathbf{x}'$ we have $\mathbf{y}$ (ground truth texture) and $\mathbf{y}'$ (generated texture), respectively. Note that the discriminator in this case, $D_{t}$, actually takes two arguments: either a real heightmap / real texture pair $(\mathbf{x}, \mathbf{y})$, or a real heightmap / generated texture pair $(\mathbf{x}, \mathbf{y}')$. Also note that for the pix2pix part of the network, we can also employ some form of pixel-wise reconstruction loss to prevent the generator from dropping modes. Therefore, we can write the training objectives for pix2pix as such:
\begin{equation}
\begin{split}
\min_{G_{t}} \ & \ \ell( D_{t}(\mathbf{x}, G_{t}(\mathbf{x}')), 1 ) + \lambda d(\mathbf{y}, G_{t}(\mathbf{x}'))\\
\min_{D_{t}} \ & \ \ell( D_{t}(\mathbf{x}, \mathbf{y}), 1 ) + \ell( D_{t}(\mathbf{x}, G_{t}(\mathbf{x}')), 0 ),
\end{split}
\label{eq:2}
\end{equation}
where $d(\cdot, \cdot)$ can be some distance function such as $L_{1}$ or $L_{2}$ loss, and $\lambda$ is a hyperparameter denoting the relative strength of the reconstruction loss. We set $\lambda = 100$ and use $L_{1}$.

\section{Experiments and Results}

\begin{figure*}
    \centering
    \includegraphics[width=0.6\textwidth]{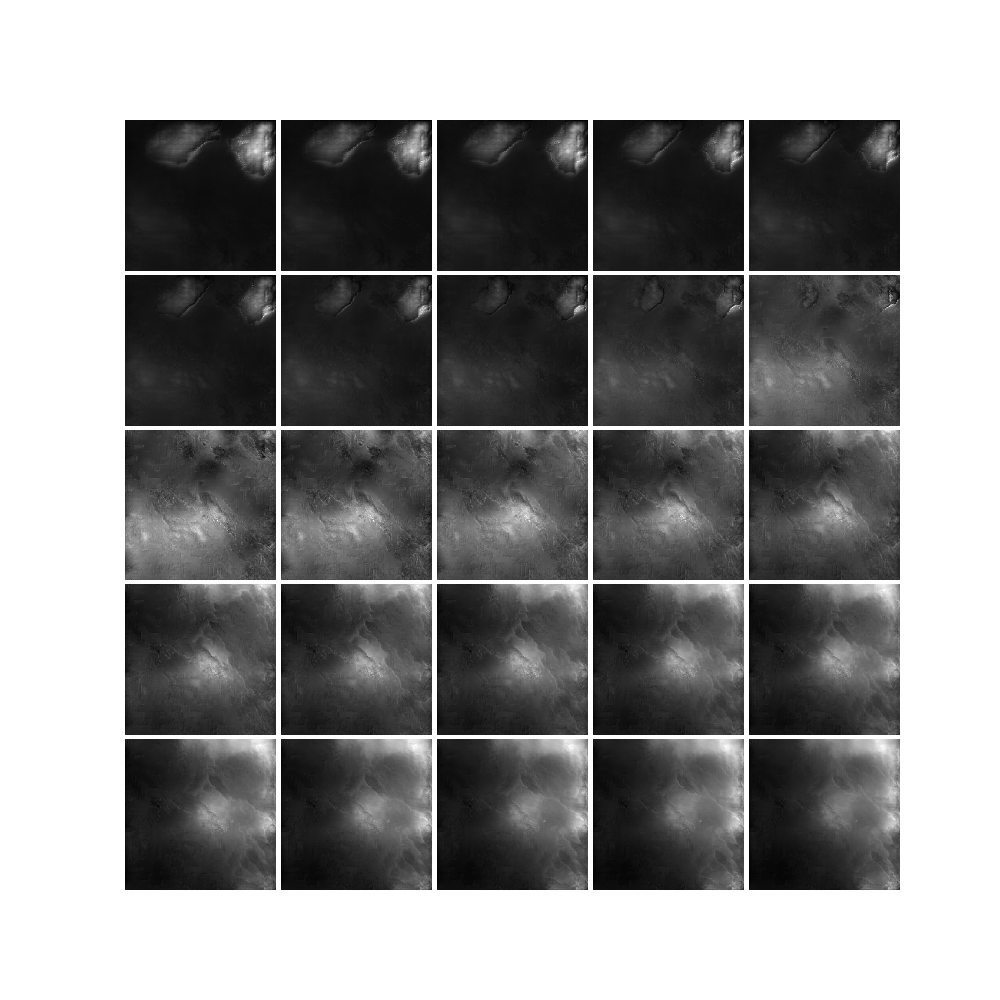}
    \caption{A linear interpolation of two heightmaps (top-left and bottom-right corner).}
    \label{fig:interp}
\end{figure*}

As a first step, we train a DCGAN which maps samples from the prior $\mathbf{z}' \sim p(\mathbf{z})$ to a generated heightmap $\mathbf{x}' = G_{h}(\mathbf{z}')$ of size 512px. While we experienced some issues with training stability we were able to generate heightmaps that were somewhat faithful to the original images. We generate two of these and illustrate a linear interpolation between the two, which is shown in Figure \ref{fig:interp}. While the interpolation is shown purely for illustrative purposes, one could imagine that if the DCGAN successfully learned representations of the different landscapes (e.g. mountains, valleys, desert, jungle) then one could find their latent representations -- through a bidirectional GAN like BiGAN \cite{bigan} or ALI \cite{ali} -- then interpolate between them and decode to control the resulting heightmap.

Apart from the aforementioned stability issues, the generated heightmaps can sometimes exhibit small-scaled artifacts, which can be seen for the first generated heightmap (top-left corner). While experimenting with deeper architectures and/or skip connections could mitigate this, one easy trick is to apply a slight blur to the final images via a Gaussian kernel convolution. This can serve to smooth out any weird artifacts generated by the DCGAN.

\begin{figure*}
    \centering
    \includegraphics[width=0.95\textwidth]{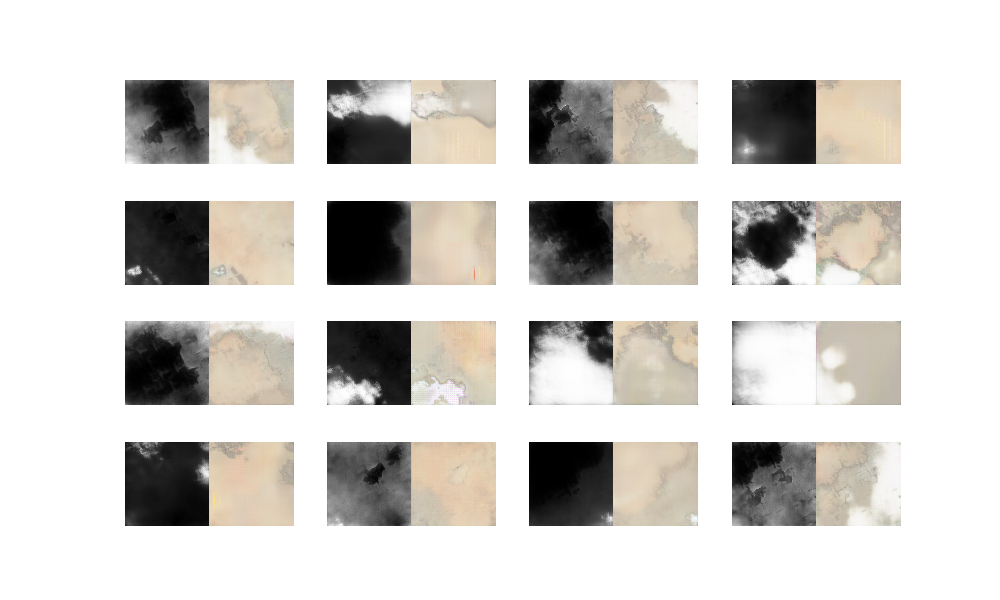}
    \caption{Heightmaps generated by the $G_{h}$ and their corresponding textures predicted by $G_{t}$. The texture GAN $G_{t}$ seems to think that high elevations are snow, despite the fact that we are generating deserts -- what is this madness?! We will just pretend they are salt flats for now.}
    \label{fig:both_gen_grid}
\end{figure*}

Figure \ref{fig:both_gen_grid} shows the a variety of heightmaps generated with the DCGAN $G_{h}$ and their corresponding translations to textures by the pix2pix GAN $G_{t}$. We can see that the pix2pix GAN has created textures that roughly `match' their corresponding heightmaps. For example, regions of relatively higher elevation in the heightmap correspond to different textures. Interestingly, parts of some of the textures are completely white; this appears to be a side-effect of not training the DCGAN and pix2pix GANs jointly.


\section{Conclusion}

In this work we have achieved a reasonable first step toward procedural generation of terrain based on real-world data. The most obvious next step would be to jointly train the DCGAN and pix2pix GANs.

A neat addition to this idea would be the addition of a segmentation pipeline to classify different parts of the terrain, e.g. biomes. This effectively serves as a layer of metadata that can be leveraged to add interesting detail in the terrain. For example, if the segmentation identifies a certain region in the generated terrain as `jungle', the 3D game engine (or renderer) can automatically populate that region with trees and plants. (This is called a `splatmap' in the computer graphics literature.)

The two-stage GAN framework that we have described here can have many applications in procedural generation outside of terrain modelling. For example, one can imagine the same scheme being applied to synthesise 3D meshes which are then textured (e.g. faces). These kinds of possibilities serve to not only promote richer entertainment experiences, but to also provide useful tools to aid content producers (e.g. 3D artists) in their work.

\section{Acknowledgements}

The authors would like to thank the developers of Theano \citep{theano}, Lasagne \citep{lasagne}, and Keras \citep{keras}. This work is partially funded by Imagia, Inc. under a MITACS Accelerate scholarship.

\begin{figure*}
    \centering
    \subfigure[Heightmap]{\label{fig:pair1}\includegraphics[width=0.4\textwidth]{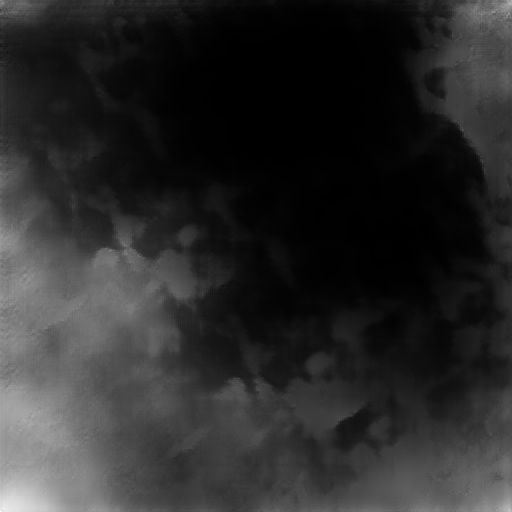}} \subfigure[Texture map]{\label{fig:pair2}\includegraphics[width=0.4\textwidth]{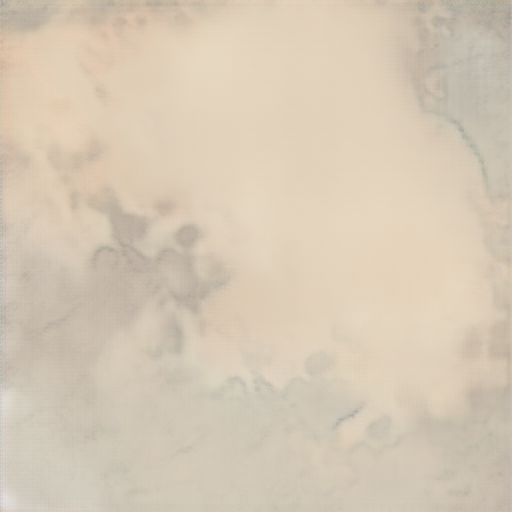}}
    \caption{Left: randomly generated heightmap, right: the corresponding texture. Both images are 512px, which corresponds to 512 square km.}
    \label{fig:pair}
\end{figure*}

\begin{figure*}
    \centering
    \includegraphics[width=0.95\textwidth]{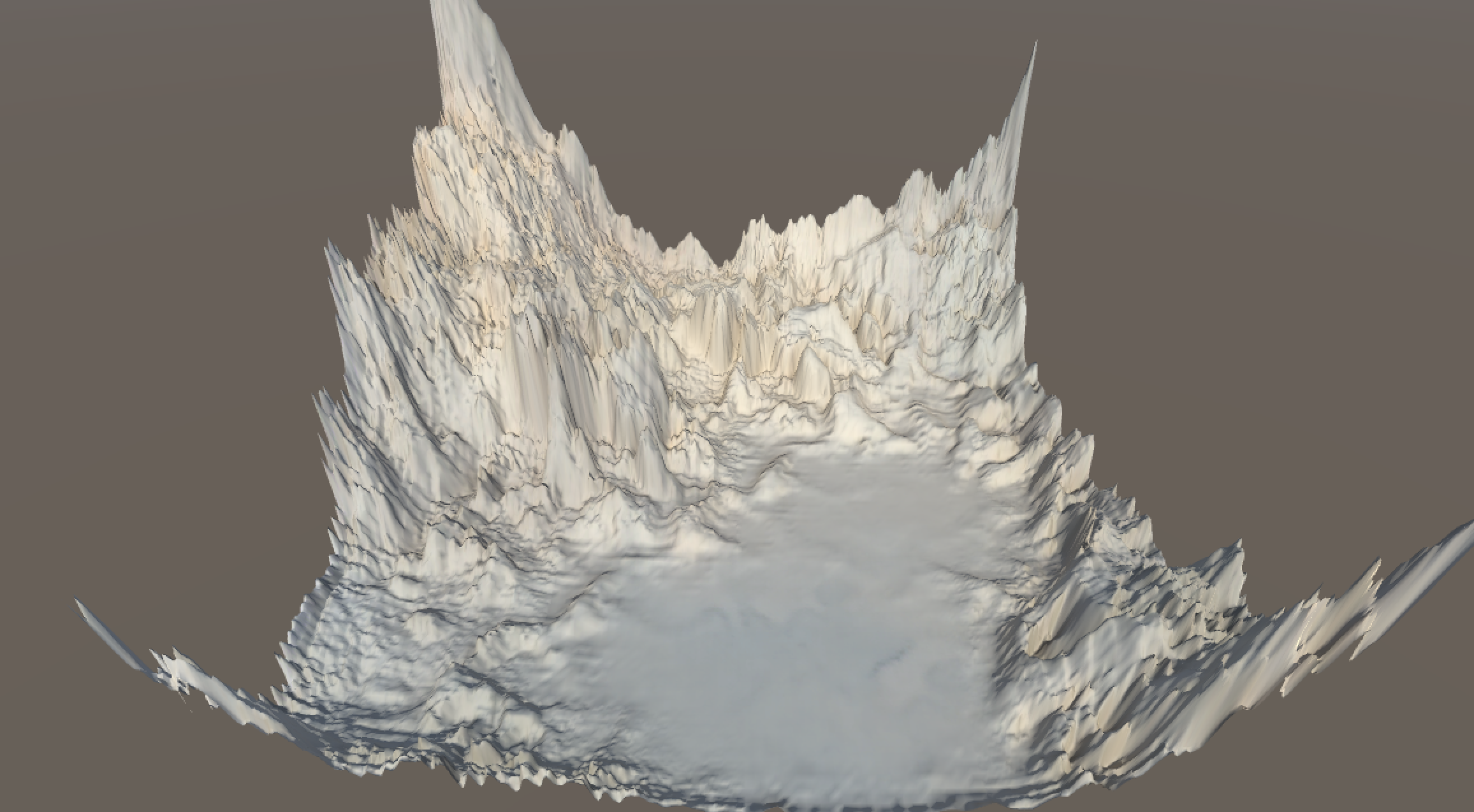}
    \caption{A rendering of one of the generated heightmap in Figure \ref{fig:pair1} in the Unity 3D game engine. A very minor Gaussian blur (of radius 0.4px) was used to smooth out artifacts prior to rendering.}
    \label{fig:hm_render_unity}
\end{figure*}

\pagebreak

\bibliography{main}
\bibliographystyle{icml2017}

\section{Supplementary material}

In this section we provide some extra information that we were unable to fit into the main section due to page restrictions.

\subsection{Dataset}

The dataset was prepared as follows. First, we downloaded a high-res heightmap and texture map of the earth, as can be found here\footnote{\url{https://eoimages.gsfc.nasa.gov/images/imagerecords/74000/74218/world.200412.3x21600x10800.jpg}} and here\footnote{\url{https://eoimages.gsfc.nasa.gov/images/imagerecords/73000/73934/gebco_08_rev_elev_21600x10800.png}}. We slide a 512px window through both images simultaneously, and only retain (heightmap, texture) pairs where the heightmap's colour composition is less than 90\% black; this so that we do not feed the GAN data that is too `trivial' to generate. Note that at this point, textures in the collection can correspond to various biomes such as jungle, desert, and arctic, and in theory, any particular heightmap could correspond to any of these biomes (which can confuse the pix2pix GAN during training). To address this, we choose a `reference texture' with our biome of interest (in our case, desert), and compute the Euclidean distance between this texture and all other textures in the collection. From this, we choose the top $M$ pairs that have the smallest distances with the reference texture, so that the final collection only contains pairs whose biome of interest is desert.

\subsection{Architectures}

Images detailing the precise architectures used for the GANs have been added in the .zip file from which the source of this \LaTeX \ can be found.

\subsection{Training}

We use the LSGAN formulation as this made training more stable. Therefore, the $\ell(\cdot, \cdot)$ in Equations \ref{eq:1} and \ref{eq:2} are binary cross-entropy, and the output activations of both discriminators $D_{h}$ and $D_{t}$ are linear instead of sigmoid.

We train both GANs using RMSProp with initial learning rates $1e^{-4}$.

\end{document}